\documentclass[sigconf]{acmart}
\AtBeginDocument{%
  \providecommand\BibTeX{{%
    \normalfont B\kern-0.5em{\scshape i\kern-0.25em b}\kern-0.8em\TeX}}}
\usepackage{verbatim}
\usepackage{graphicx}
\usepackage{algorithm}
\usepackage{algpseudocode}
\usepackage{float}
\usepackage{xcolor}
\usepackage{makecell}

\copyrightyear{2020}
\acmYear{2020}
\setcopyright{acmcopyright}\acmConference[CHI PLAY '20]{Proceedings of the Annual Symposium on Computer-Human Interaction in Play}{November 2--4, 2020}{Virtual Event, Canada}
\acmBooktitle{Proceedings of the Annual Symposium on Computer-Human Interaction in Play (CHI PLAY '20), November 2--4, 2020, Virtual Event, Canada}
\acmPrice{15.00}
\acmDOI{10.1145/3410404.3414235}
\acmISBN{978-1-4503-8074-4/20/11}

\begin{document}

%%
%% The "title" command has an optional parameter,
%% allowing the author to define a "short title" to be used in page headers.
\title{Predicting Game Difficulty and Churn Without Players}

%%
%% The "author" command and its associated commands are used to define
%% the authors and their affiliations.
%% Of note is the shared affiliation of the first two authors, and the
%% "authornote" and "authornotemark" commands
%% used to denote shared contribution to the research.
\author{Shaghayegh Roohi}
\email{shaghayegh.roohi@aalto.fi}
\affiliation{%
	\institution{Aalto University}
	\city{Espoo}
	\country{Finland}
}

\author{Asko Relas}
\email{asko.relas@rovio.com}
\affiliation{%
	\institution{Rovio Entertainment}
	\city{Espoo}
	\country{Finland}
}

\author{Jari Takatalo}
\email{jari.takatalo@rovio.com}
\affiliation{%
	\institution{Rovio Entertainment}
	\city{Espoo}
	\country{Finland}
}

\author{Henri Heiskanen}
\email{henri.heiskanen@rovio.com}
\affiliation{%
	\institution{Rovio Entertainment}
	\city{Espoo}
	\country{Finland}
}

\author{Perttu H\"am\"al\"ainen}
\email{perttu.hamalainen@aalto.fi}
\affiliation{%
	\institution{Aalto University}
	\city{Espoo}
	\country{Finland}
}
%%
%% By default, the full list of authors will be used in the page
%% headers. Often, this list is too long, and will overlap
%% other information printed in the page headers. This command allows
%% the author to define a more concise list
%% of authors' names for this purpose.
\renewcommand{\shortauthors}{Trovato and Tobin, et al.}

%%
%% The abstract is a short summary of the work to be presented in the
%% article.
%!TEX root = main.tex
\begin{abstract}
	We propose a novel simulation model that is able to predict the per-level churn and pass rates of Angry Birds Dream Blast, a popular mobile free-to-play game. Our primary contribution is to combine AI gameplay using Deep Reinforcement Learning (DRL) with a simulation of how
	the player population evolves over the levels. The AI players predict level difficulty, which is used to drive a player
	population model with simulated skill, persistence, and boredom. This allows us to model, e.g., how less persistent and skilled players are
	more sensitive to high difficulty, and how such players churn early, which makes the player population and the relation
	between difficulty and churn evolve level by level. Our work demonstrates that player behavior predictions produced by DRL gameplay can be significantly improved by even a very simple population-level simulation of individual player differences, without requiring costly retraining of agents or collecting new DRL gameplay data for each simulated player. %This is important as training the DRL agents can be very slow (60 hours per game level in our case). %As training the DRL agents can be very slow, this is a foundational result that can empower future research and applications of player modeling and simulation-based game testing. 

%To fit the model to observed pass and churn rates from real players, we optimize three types of parameters: 1) linear regression weights mapping AI gameplay features to difficulty, 2) initial distribution of player traits, and 3) the parameters of population simulation. %The results clearly improve a baseline of predicting churn and pass rate from the AI gameplay alone, without the population simulation.
\end{abstract}

%%
%% The code below is generated by the tool at http://dl.acm.org/ccs.cfm.
%% Please copy and paste the code instead of the example below.
%%


%%
%% Keywords. The author(s) should pick words that accurately describe
%% the work being presented. Separate the keywords with commas.

%% A "teaser" image appears between the author and affiliation
%% information and the body of the document, and typically spans the
%% page.
\begin{teaserfigure}%
	\centering
	\includegraphics[width=\linewidth]{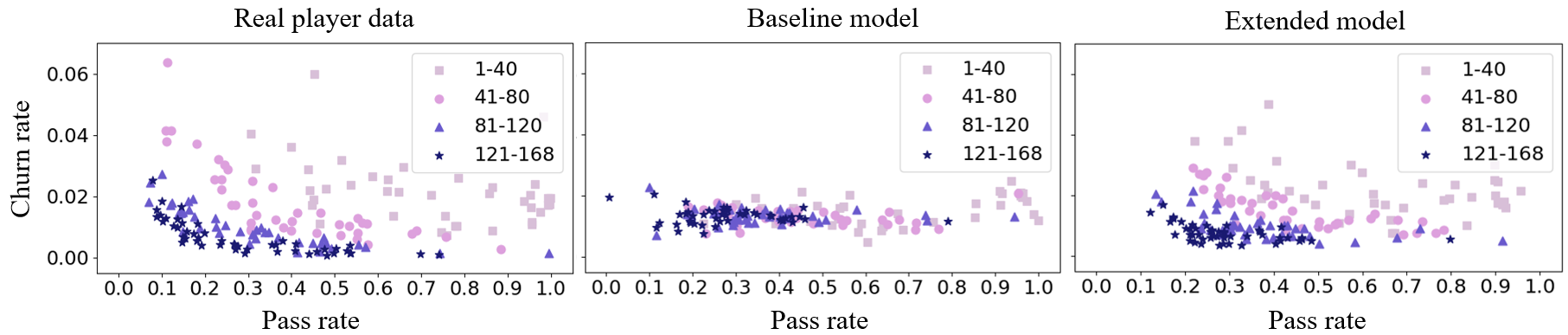}
	\caption{Scatter plots depicting the relation of pass rate (a measure of level difficulty) and churn rate over 168 game levels of Angry Birds Dream Blast, in both real player data and our simulations. Here, churn is defined as not playing for 7 days. The colors denote level numbers. The baseline simulation model predicts pass rate and churn directly from AI gameplay. Our proposed extended model augments this with a simulation of how the player population evolves over the levels.}
	\label{fig:3_plots}
	\vspace{-0.1cm}
\end{teaserfigure}

%%
%% This command processes the author and affiliation and title
%% information and builds the first part of the formatted document.
\maketitle

%!TEX root = main.tex
\section{Introduction}
% say something about the free-to-play games.%
One of the primary difficulties of game design and development is that player behavior is hard to predict. This leads to an iterative design process of prototyping and testing, which is slow and expensive. Ideally, research should produce models and tools that allow evaluating the effect of design decisions early on, before committing resources to real-life game testing. This is one of the foundational motivations of player and user modeling  \cite{oulasvirta2018computational,pedersen2010modeling,yannakakis2013player}.

Better models and tools are in particular needed for predicting and optimizing business critical behavior such as churn, i.e., a player quitting the game and not coming back to it. Churn matters as many modern games accumulate their revenue gradually from in-game advertisements and purchases, instead of single up-front fee. To prevent churn, free-to-play game companies engage in extensive data-driven A/B-testing and optimization of game levels. For example, one would like to identify and modify levels with high churn rate. However, instead of deploying different game level versions to real players and seeing what happens, it would be desirable to have predictive models of churn that allow designers to conduct initial testing and prototyping \emph{in silico}.

%Many modern games accumulate their revenue gradually from in-game advertisements and purchases, instead of single up-front fee. Therefore, it is preferable to have players playing for a long time, instead of \emph{churning}, i.e., leaving the game \cite{kristensen2019combining}. To prevent churn, free-to-play game companies engage in extensive data-driven A/B-testing and optimization of game levels. For example, one would like to identify and modify levels with high churn rate. However, instead of deploying different game level versions to real players and seeing what happens, it would be desirable to have predictive models of churn that allow designers to conduct initial testing and prototyping \emph{in silico}.

An emerging possibility is to use simulated game-playing agents instead of human playtesters \cite{pedersen2010modeling,ariyurek2019automated}, at a fraction of the time and cost. However, present models are limited, e.g., by the need to train models with large amounts of real player data in a way that does not necessarily generalize to new game content \cite{gudmundsson2018human}. Models and algorithms that can play games without a large dataset of ground truth behaviors are often not validated to give realistic predictions of player behavior and experience \cite{roohi2018review}. %\textcolor{red}{TODO: what else can one say about limitations and the current knowledge gap?, e.g., in light of ariyurek 2019} \textcolor{blue}{According to ariyurek et al. \cite{ariyurek2019automated}, automatically defining test scenarios to find the game defects is another limitation. We can mention that creating AI agents that shows various game playing strategies needs lots of manually-designed heuristics and utility functions which is a hard task for complex games and might end up into an incomplete range of playing styles\cite{holmgard2018automated,guerrero2018using,mugrai2019automated}.}

This paper makes the following contributions to player modeling and simulated game testing:

\begin{enumerate}
\item We present new data indicating a relation between game level difficulty (measured as pass rate) and churn rate, shown in Figure 1. The data is from 168 levels and 95266 players of Angry Birds Dream Blast, a successful free-to-play mobile game from Rovio Entertainment. 
\item We propose a novel simulation model that predicts the observed relation of pass and churn rates. Our key innovation is to combine Deep Reinforcement Learning (DRL) game playing agents with a simulation of how a player population with simple computational models of skill, persistence, and boredom evolves over the levels of a game as some players churn.
\end{enumerate}

Previous work has investigated DRL gameplay \cite{justesen2019deep} and player population simulation \cite{reguera2019physics}, but not combined the two. Previous work has also predicted pass rate \cite{gudmundsson2018human} and churn \cite{perianez2016churn,bonometti2019modelling,yang2019mining}, and found a relation between game success rate and engagement \cite{lomas2013optimizing,lomas2017difficulty}, but has not modeled how the two interrelate in a dynamic manner over a game's levels. 

\section{Related Work}\label{sec:related_work}
\subsection{Churn prediction}
Churn means inactivity duration, and its precise definition varies from a few days to completely quitting a game.
Churn prediction empowers game companies to engage players who are likely to churn, and hence increase the success probability of the game. Churn prediction has been approached through survival ensembles and Cox regression to estimate the likelihood of not being churned and survive after a specific amount of time \cite{perianez2016churn, bertens2017games}. Bertens et al. \cite{bertens2017games} inspected both survival ensembles and Cox regression for using in-game features such as player logins, play time, purchases, and level ups in churn prediction.

Other methods address churn prediction as a classification problem. Bonometti et al. \cite{bonometti2019modelling} employ deep neural networks to jointly estimate survival time and churn probability by modeling early interactions between players and the game, using the metrics like play time, time difference between play sessions, and in-game activity type and diversity. Yang et al. \cite{yang2019mining} study the regularity of the time that long-term players spend in the game to perform the binary classification into churned and not-churned .

In the context of churn classification, some works use aggregated data \cite{yang2019mining}, and others exploit temporal data \cite{kim2017churn, lee2018game} instead. Bonometti et al. \cite{bonometti2019modelling} compared aggregated and temporal data, and find that models with temporal data outperform the other kind. Kristensen et al. \cite{kristensen2019combining} propose using stacked LSTM networks with a combination of aggregated and time-series data as their inputs.

In contrast to the work above, we do not predict the probability that a particular player churns. Instead, we predict the average churn ratio per game level, with the goal of helping game designers identify problematic levels before they are deployed to players. We employ game-playing AI agents in producing the input features of the predictor.
 
\subsection{Game-playing Agents}
Several studies have investigated game-playing agents in video games. Some of these studies have focused on Monte Carlo tree search (MCTS) \cite{poromaa2017crushing, holmgard2018automated, ariyurek2020enhancing}, and others have exploited advances in deep learning methods \cite{mnih2013playing, vinyals2017starcraft, lample2017playing, gudmundsson2018human, kamaldinov2019deep}. For in-depth overviews of recent research on the topics, we refer the reader to Shao et al. \cite{shao2019survey} and Justesen et al. \cite{justesen2019deep}. 

Game-playing agents are beneficial in playtesting by reducing costs and the need for human playtesters \cite{ariyurek2019automated}. AI agents have been found useful in finding bugs \cite{de2017ai}, game parameter tuning \cite{isaksen2015exploring}, and predicting level difficulty and average player pass rate \cite{poromaa2017crushing, gudmundsson2018human}.

Agents that are used in playtesting need to be reasonably human-like. Zhao el al. \cite{zhao2019winning} defines human-likeness as a balance between skill and playing style. Ariyurek et al. \cite{ariyurek2019automated} compare MCTS and human-like AI agents in finding bugs. One can train human-like game-playing agents simply through deep neural networks and imitation learning \cite{gudmundsson2018human}, although this requires large amounts of human gameplay data and does not necessarily generalize to new content. For instance, if there are no human reference actions to imitate for a new type of a game puzzle, an imitation learning agent's behavior is undefined. 

An alternative to imitation learning is provided by the computational rationality paradigm, which views human behavior as emerging from optimization of rewards received through taking actions, albeit with limited computational capabilities \cite{gershman2015computational}. The rewards can be either extrinsic ones such as a game score, or intrinsic ones defined by computational motivation and emotion models \cite{roohi2018review}. A common way to implement such reward optimization is through MCTS, which requires no lengthy training, but incurs a high run-time computing cost. For instance, Poromaa et al. \cite{poromaa2017crushing} uses a MCTS agent's performance to predict the average human pass rate. An alternative is Deep Reinforcement Learning (DRL), which requires a lengthy and potentially unstable training process, but can yield computationally lightweight neural network agents. DRL is also better suited for continuous state and action spaces, which are needed for embodied physically simulated players. DRL has recently been shown to enable user simulation including the biomechanics of the human body, predicting the perceived shoulder fatigue of pointing movements \cite{cheema2020predicting}.

%Stefan et al. \cite{gudmundsson2018human}, uses convolutional neural network to learn human players actions in visual game states in a supervised learning way. Moreover, they map statistics gathered from the learned agent's gameplay to the difficulty of levels in a tile-matching game. 

In addition to imitation learning and computational rationality approaches, some works implement heuristic agents based on the common strategies that players take during playing the game \cite{de2017ai, holmgard2018automated}. Isaksen et al. \cite{isaksen2015exploring} optimize game parameters to reach a specific level of difficulty based on a heuristic AI player' score. Silva et al.  \cite{de2017ai} suggest how analysis of AI gameplay can be helpful in finding defects in the rule set of a board game. They explored various aspects of the game by matchups between game-specific agents, each with different playing style. More recently, Holmgard et al. \cite{holmgard2018automated} evaluate the effect of enhancing MCTS agents with player type heuristics on the playablity of multiple dungeon maps.

In this paper, we focus on DRL agents that require no hand-coded heuristics or human player data for learning to play a game. Human data is only required for fitting a simple "sim2real" model --- e.g., linear regression --- that maps simulation results to human-like churn and pass rates. %did not use human data or heuristic-based agents to play the game, but an intrinsically-motivated agent was used to play the game. This type of agents are self-supervised agents, which play the game for their own desires, and do not need manually-tuned reward function.

\subsection{Understanding Game Difficulty}
Our work is motivated by the relation of level pass rate and churn rate observed in our data. Pass rate is a measure of level difficulty, and the effect of difficulty on player experience is a topic with a considerable body of research. One of the most prevalent models in game design literature is the flow channel  \cite{takatalo2010presence,schell2008art}, according to which the ideal game is not too easy and not too hard, to keep the player in the sweet spot between boredom and anxiety. A similar view is provided by intrinsic motivation theories: Self-determination theory posits the basic psychological need to feel competent \cite{ryan2006motivational}, which a too difficult game can obviously destroy. On the other hand, research has also identified a basic need for novelty of stimuli and experiences, central to experiencing curious interest \cite{sheldon2001satisfying,kashdan2009curiosity,lomas2017difficulty}. Competence and curiosity interact, as one way to keep a game easy is to avoid introducing new types of challenges the player has to learn, which violates the need for novelty. In practice, successful games seem to optimize for both competence and curiosity, using a carefully tuned difficulty progression that allows experiencing competence, and periodically resets back to a low difficulty when introducing new mechanics and challenges to maintain novelty \cite{linehan2014learning}. Recent work on mathematically optimal learning, e.g., in artificial neurons, also posits a flow-channel style difficulty model \cite{wilson2019eighty}, identifying 85\% success rate as optimal. %Looking at popular puzzle games where success could be defined as passing a level, the 85\% rule seems to fit reality fairly well. At least during the initial on-boarding phase of games, player success rate seems to be closer to 100\% than 50\%.

Naturally, preferences for difficulty vary between players and games. For example, recent research has provided explanations on why frustratingly difficult games can sometimes be enjoyable \cite{petralito2017good}. However, our data is more in line with Lomas et al. \cite{lomas2013optimizing}, who found lower difficulty to yield the highest engagement in their educational game. In a follow-up study, Lomas et al. \cite{lomas2017difficulty} found moderate difficulty as most motivating when players could select their opponent's skill ranking. However, the easiest games were most motivating when difficulty was determined randomly.

%!TEX root = main.tex
\section{Data: Angry Birds Dream Blast}\label{sec:game_discription}
\begin{figure}[h]
	\centering
	\includegraphics[scale=0.7]{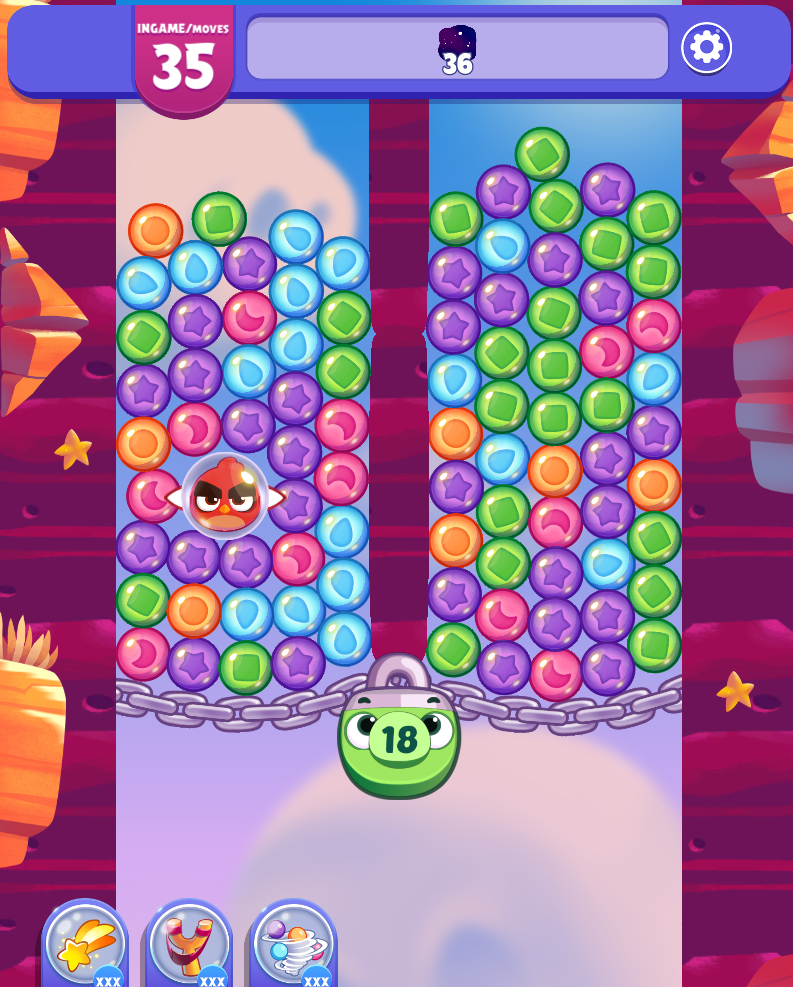}
	\caption{Screenshot of Angry Birds Dream Blast.}	
	\label{fig:db_screenshot}
\end{figure}

This paper focuses on Angry Birds Dream Blast \cite{DB}, a successful free-to-play mobile game. Angry Birds Dream Blast is a non-deterministic, physics-based match-3 game with a limited number of moves per level. The player pops adjacent bubbles with similar colors. Figure \ref{fig:db_screenshot} shows a screenshot of the game. Beyond simple bubble popping, different types of boosters can be produced by combining bubbles. These boosters are used to pop a large quantity of bubbles in rows and columns. Boosters are required to remove some objects in the game. Additionally, there are locks in the game that can be unlocked by collecting bubbles or other objects. Players can also acquire power-ups and extra moves through in-app purchases. However, this is a feature we did not enable for the AI agents. 

For our study, we utilized a dataset of per-level pass and churn rates from a total of 95266 players. Per-level churn rate, in range 0...1, was defined as the portion of players who stopped playing for at least 7 days after trying the level at least once. After the 7 day period, some players may return to the game, but churn measurement time needs to be limited in practice, e.g., for A/B testing of game content updates. Pass rate, also in range 0...1, was computed as the mean of 1 divided by the number of attempts players required to pass each level. 

%Overall, our data indicates a strong relation between pass rate and churn. Although the overall correlation of pass rate and churn rate is low (Spearman's $r=-0.144$), the correlation is strong when computing churn rate only from players who failed to pass a level (Spearman's $r=-0.0.586$). 
Figure \ref{fig:3_plots} indicates a relation between real player pass and churn rates. Although the overall correlation is small (Spearman's $r=-0.144$), the correlation is large when computing churn rate of a level as the portion of players who churn without completing the level (Spearman's $r=-0.586$). In other words, players who churn without completing a level are more likely to do so in the more difficult levels. Beyond a simple correlation, Figure \ref{fig:3_plots} also suggests that a low pass rate causes less churn in the later levels. It is this dynamic we set out to explain in this paper.

It should be noted that Angry Birds Dream Blast is a live game that is constantly evolving. Our data was collected from a version that does not have all the mechanics of the latest game version. %Also note that since pass and churn rates are critical business data, we can only plot the variables without exact numbers. However, this does not affect our analysis of how the variables interrelate. 

%UNCOMMENT THE ABOVE IF WE NEED TO REMOVE THE NUMBERS

%!TEX root = main.tex
\section{Method}\label{sec:method}
\begin{figure*}[!t]
	\centering
	\includegraphics[width=0.6\textwidth]{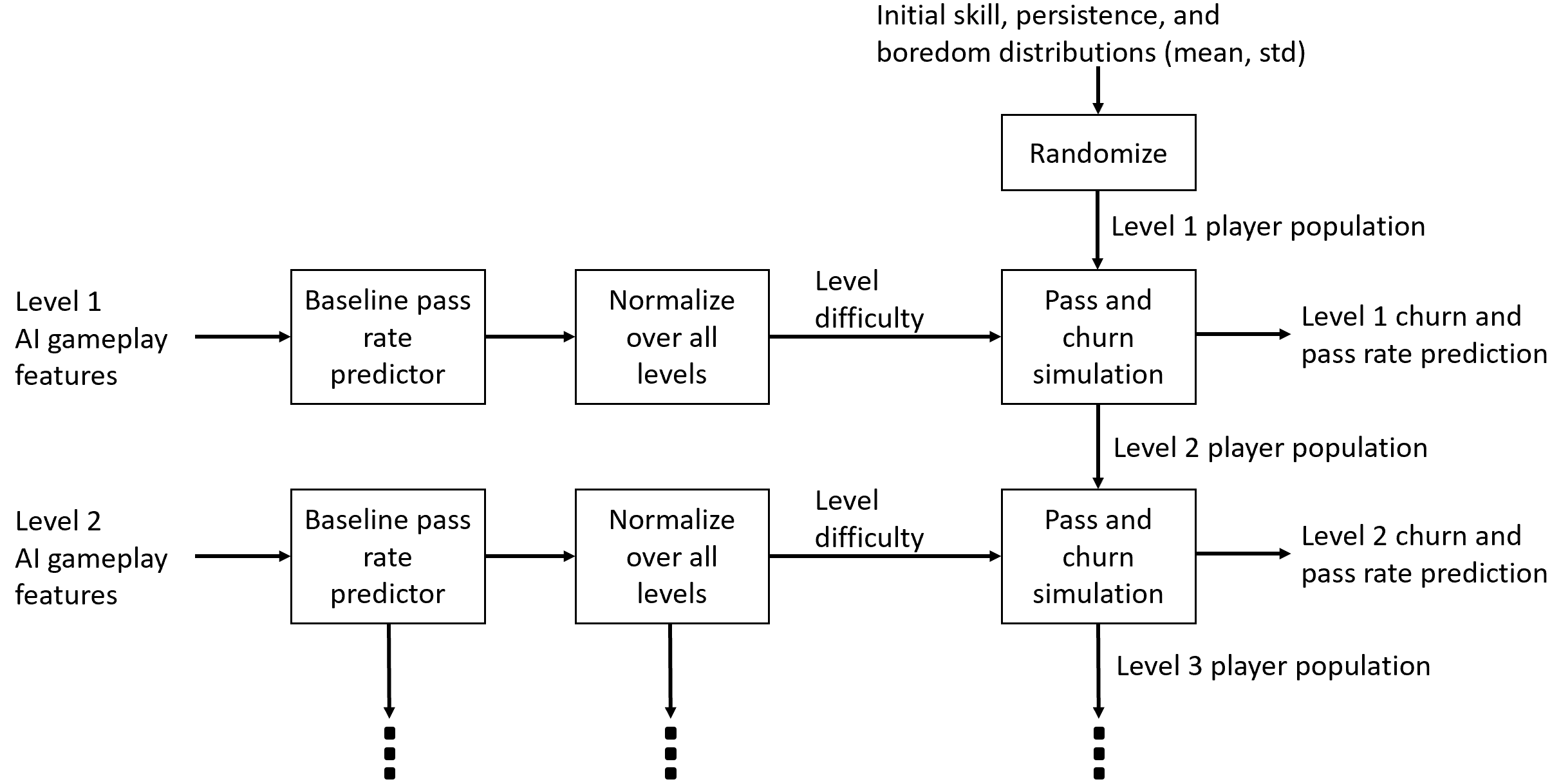}
	\caption{Overview of our extended simulation model, showing how AI gameplay data of each level results in a level difficulty estimate, which is fed to pass and churn simulation, together with player population parameters. The simulation outputs both the churn and pass rate predictions of a level, and the player population for the next level.}
	\label{fig:mc_algorithm}
\end{figure*}

We implement and compare two churn and pass rate prediction models: 

\begin{enumerate}
\item A baseline regression model that directly predicts churn and pass rate from AI gameplay, motivated by earlier work on predicting game difficulty by AI gameplay \cite{gudmundsson2018human} and the relationship between pass rate and churn observed in our data in Figure \ref{fig:3_plots}, echoed by previous research on game difficulty and engagement \cite{lomas2013optimizing,lomas2017difficulty}.

\item The extended model in Figure \ref{fig:mc_algorithm} that combines both AI gameplay and player population evolution over game levels.
\end{enumerate}

The per-level pass and churn simulation of the extended model takes in both a level difficulty estimate based on the AI gameplay, and a player population. The simulation outputs both the churn and pass rate predictions of a level, and the new player population for the next level, with churned players removed from the population.  In determining the passed and churned players, we employ simple computational models of the following psychological and learning phenomena:

\begin{itemize}
\item A player passes a level if the level difficulty is lower than the player's skill. 
\item A player churns if taking more attempts to pass a level than the player's persistence. This allows us to model how less persistent players churn earlier, leading to later levels being played by more persistent players who are less sensitive to churn because of high difficulty.
\item Even if a player passes a level, they may churn with some random probability proportional to the player's tendency to get bored. 
\item Passing a level gets gradually easier over multiple attempts, as the player learns from their mistakes.
\end{itemize}

Below, we first describe our AI game playing agent, then explain how the AI gameplay data is used for the baseline model. Finally, we detail the extended model of Figure \ref{fig:mc_algorithm}, including the pass and churn simulation (Algorithm \ref{algo:simulation}).

\subsection{AI Gameplay}\label{sec:drl}
Our Deep Reinforcement Learning AI agents are implemented using the Unity ML-agents framework \cite{juliani2018unity} and Proximal Policy Optimization (PPO) \cite{schulman2017proximal} algorithm. 

In Reinforcement Learning, an agent observes system state and takes an action, which leads it to receives a new observation and a reward. PPO optimizes the policy --- a distribution of actions conditional on state observations, implemented as a neural network --- to maximize the agent's expected cumulative future rewards. 

To make game playing emerge with PPO, one needs to frame the game as a RL problem by defining the state observations, actions, and rewards. Furthermore, one needs to define the neural network architecture and a number of algorithm parameters. Our choices are detailed below.

\subsubsection{State observations} A combination of visual and numerical vector observation were used. The visual observation is a $84 \times 84 \times 3$ RGB screenshot of the game, and the vector observation contains the number of moves left, the types and numbers of the remaining level goals (e.g., collecting a specific number of bubbles), the numbers and types of the objects needed to unlock the visible locks, and game camera position.

\subsubsection{Action space} We use discrete action space, discretizing the game screen into $32 \times 32$ possible points that the agent can click/tap. A continuous action space was also tested, but did not produce as good results. 

\subsubsection{Reward function} The reward function combines both extrinsic rewards and an intrinsic reward generated by the self-supervised curiosity model of Pathak et al. \cite{pathak2017curiosity} implemented in Unity ML-agents.

The extrinsic reward is computed as a sum of reward components: win bonus or lose penalty, cleared goals ratio, progress percentage in opening locks, a small constant negative reward that penalizes the agent for using more moves than needed, and a click reward. The click reward is calculated as $r_{distance} = c_0 \exp\left(-\frac{d}{c_1}\right)$, where $d$ is the distance to the closest available match, and $c_0$ and $c_1$ are tuning parameters. During the initial learning exploration where the agent has no idea where to click, the click reward provides extra guidance towards only clicking on valid targets.

%The agent strives to increase the intrinsic reward of curiosity which is computed as the prediction error in the forward dynamic model. Intrinsic rewards help us to train the agent without manually customized reward function, but in a self-supervision way \cite{pathak2017curiosity}.

\subsubsection{Training}\label{sec.training} We train the agent separately on each level for 5 million PPO iterations, which takes on average 60 hours per game level. Multiple training runs were conducted in parallel using cloud computing (Amazon Web Services \cite{AWS}). Each run was performed on an AWS instance with a 16-core 3.4 GHz Intel Xeon CPU, and used approximately 10 Gb RAM.

We provide the agent with %higher amounts of available moves than the actual game,
4x more moves per level than available for human players, so that the positive rewards for completing a level are less sparse. Too sparse rewards are a common reason for RL agents not learning. The extra moves also help if the level difficulty is too high and the number of moves is not sufficient even for human players.

\subsubsection{Neural network}
\begin{figure}[h]
	\centering
	\includegraphics[width=\linewidth]{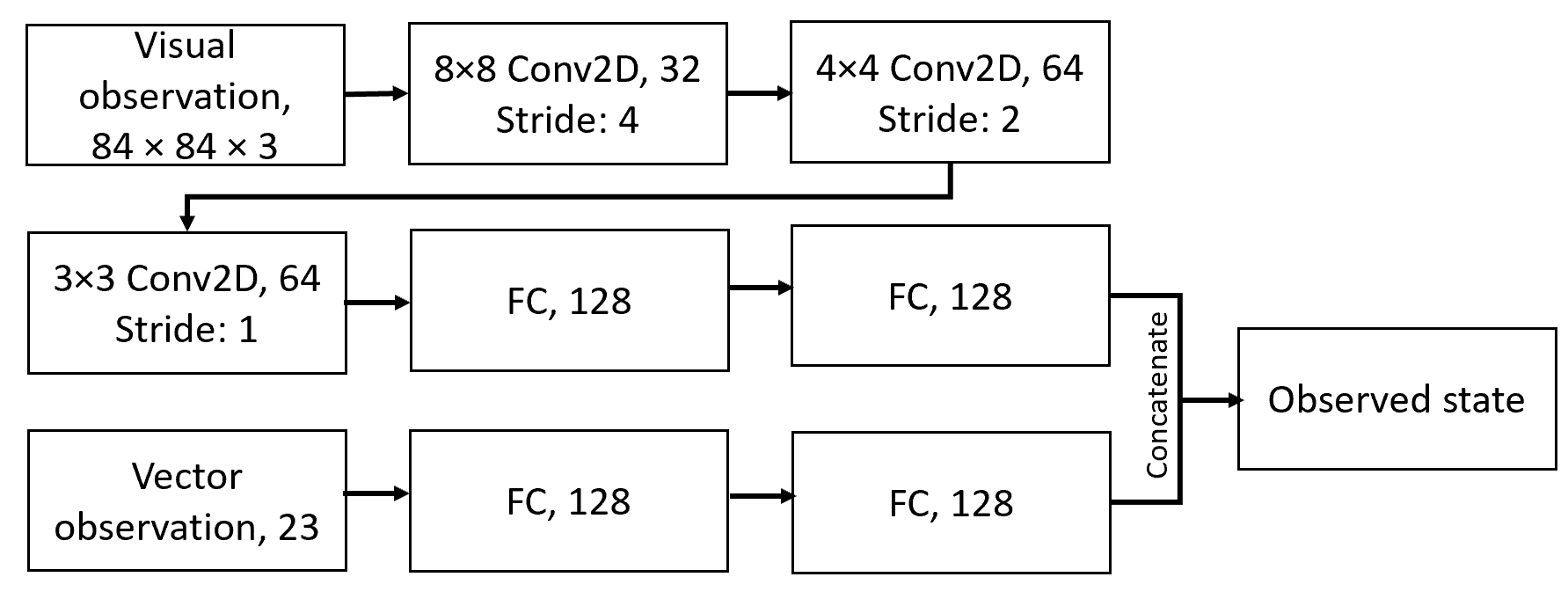}
	\caption{Neural network architecture of the observation encoder, combining both convolutional and fully connected (FC) layers.}
	\label{fig:neural_network}
\end{figure}

Visual observations are encoded using a convolutional neural network, with the result concatenated with the other observations as shown in Figure \ref{fig:neural_network}. This is one of the default settings provided in Unity ML-agents, based on the architecture originally proposed by Mnih et al. \cite{mnih2015human}.   
\subsubsection{PPO and Unity ML parameters}
The iteration experience budget of PPO agent is 10240 game moves. The agent collects experiences in episodes with a large time horizon of 1024, i.e., each episode is played until the level is completed or the agent runs out of moves. We use optimization batch size 1024, decaying learning rate initialized at 0.0003, PPO clipping parameter of 0.2, PPO entropy coefficient 0.005, discount factor $\gamma=0.99$, Generalized Advantage Estimation $\lambda=0.95$, and curiosity reward coefficient 0.02.

The parameters are explained in detail in the original PPO paper \cite{schulman2017proximal} and Unity ML-agents \cite{juliani2018unity}. The maximum number of moves per level multiplier and the curiosity reward coefficient were tuned manually. Other parameters use the Unity ML-agents default values.

\subsection{Baseline model}
The baseline model predicts both churn and pass rate using a simple least squares linear regression of the form $\mathbf{x}^T\mathbf{w}+b$, where $\mathbf{x}$ is a feature vector for the game level of interest, $\mathbf{w}$ is a vector of regression weights, and $b$ is a bias term. The feature vector comprises the mean, standard deviation, min, max, and the 5th, 10th, 25th, 50th, and 75th percentiles of the cleared goals percentage while using the moves available for human players, and the mean, standard deviation, and the 5th, 10th, and 20th percentiles of the amount of moves left when passing the level, and the mean and standard deviation of the AI agent's pass rate while using the same number of moves as human players. In total, this yields 16 features.

\begin{figure}[h]
	\centering
	\includegraphics[scale=0.32]{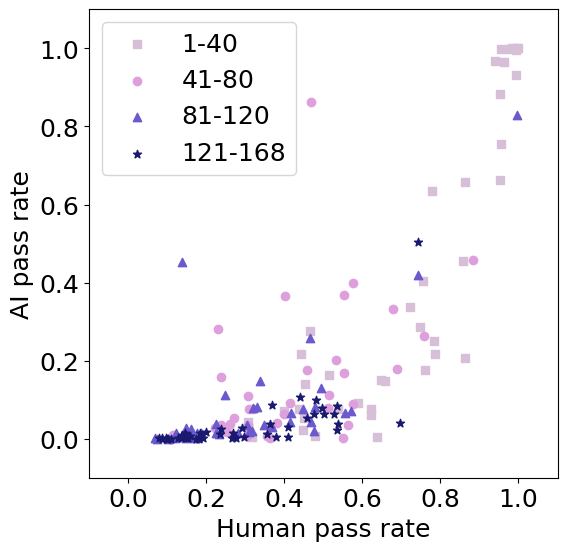}
	\caption{Relation between AI and human pass rate. Scatter plot colors correspond to game level indices.}	
	\label{fig:AI_human_passrate}
\end{figure}

Figure \ref{fig:AI_human_passrate} shows the relation between AI and human pass rate over levels. AI pass rate alone is not a great predictor of human pass rate, especially in the later game levels, where AI is only rarely able to pass the levels with the number of moves allowed for human players. As explained in section \ref{sec.training}, the AI can use 4x more moves than human players. However, other AI gameplay statistics like cleared goals percentage and moves left ratio help in the prediction.

%Figure \ref{fig:baseline_algorithm} shows two steps of the baseline method. First, agent's gameplay statistics including mean, standard deviation, min, max, and different percentiles of pass rate, moves left ratio, and cleared goals rate are collected for each game level. These 16 gathered input features are used to predict the average of players pass rate at each game level via a simple linear regression (Equation \ref{eq:baseline_reg1}). Second, another linear regression model (Equation\ref{eq:baseline_reg2}) maps the predicted pass rate into the average players churn ratio.
%
%The reason behind squared root of pass rate in equation \ref{eq:baseline_reg2} is that, we aim to model the u-curve shape (higher churn ratio when the level is too easy or difficult) between pass rate and churn ratio.
%\begin{equation}\label{eq:baseline_reg1}
%	\mathbf{p} = \mathcal{X}\mathbf{w}_1+\mathbf{b}_1,
%\end{equation}
%\begin{equation}\label{eq:baseline_reg2}
%\mathbf{c} = w_2\mathbf{p} +
%w_3\mathbf{p}^2+\mathbf{b}_2,
%\end{equation}
%where $\mathcal{X}$ is the agent gameplay statistics. $\mathbf{p}$ and $\mathbf{c}$ correspond to predicted pass rate and churn ratio.
%\begin{figure}[H]
%	\centering
%	\includegraphics[width=\linewidth]{figures/baseline_algorithm.png}
%	\caption{Baseline algorithm}
%	\label{fig:baseline_algorithm}
%\end{figure}

\subsection{Extended Model}
A basic problem of the baseline model is that it doesn't take into account how the player population varies over the levels. Because the gameplay RL formulation does not include individual differences of players, and because the AI is trained on all levels, the difficulty predicted by the AI is an average measure that cannot map well to the difficulty experienced by players with different skills and personalities. Individual differences could be implemented as reward function terms determined by intrinsic motivation and emotion models \cite{roohi2018review}, but this would incur a considerable computational cost as each different agent would require its own lengthy training process. This is why our extended model simulates individual differences using a second, computationally much more simple population-level simulation model. 

As shown in Figure \ref{fig:mc_algorithm}, our extended model uses the baseline pass rate predictor as a building block, in conjunction with a pass and churn simulation block that also takes in the current player population, and outputs the population remaining after some players have churned. This way, the distribution of player skills and other attributes is allowed to evolve throughout the game's levels. The DRL gameplay data needs to only be collected once, after which the population simulation can be run with different parameters with a very low computing cost. One simulation through all 168 levels takes less than a second on a standard desktop computer.
%It should be noted that both the DRL agents and the pass and churn simulation simulate gameplay, but the latter simulation is on a very abstract level. We simply draw a random number and map that to the outcome of a player attempting to pass a level. This is orders of magnitude more simple than the DRL gameplay which runs the actual game. 

\subsubsection{Pass and Churn Simulation Algorithm}
The pass and churn simulation is detailed in Algorithm \ref{algo:simulation}. We simulate a population of 2000 players, each player described by a tuple of real-valued attributes for skill, persistence, and boredom tendency. The initial player population's attributes are sampled from a normal distribution. 

%To implement the above, a player is described by a tuple of real-valued attributes for skill, persistence, and boredom tendency. The player population is described as a collection of such tuples. 

For each level, we draw two random numbers $s,t$ from normal distributions with means determined by player skill and persistence attributes, and standard deviations defined by the parameters $\alpha,\beta$ (Algorithm \ref{algo:simulation} lines \ref{line:s}-\ref{line:t}). Each simulated player attempts a level until passed or churned (line \ref{line:while}). If the random draw $s$ exceeds the level difficulty, the player passes (line \ref{line:ifpassed}). Otherwise, the player churns if the number of attempts exceeds the random draw $t$ (line \ref{line:ifchurn}). Additionally, a player also churns if passing the level but becoming bored, which is determined by the third random draw $b$ based on the boredom attribute (lines \ref{line:b}-\ref{line:ifbored}). For each failed attempt, the players are also learning from their mistakes, simulated as incrementing $s$ (line \ref{line:learning}). At the end, we return the population to its original size by replicating randomly selected players (line \ref{line:resample}). Because of the random selection, this replication does not affect the distribution of individual differences, and it prevents population depletion that would lead to inaccuracies in simulating later levels.

Note that we only simulate per-level learning, instead of learning over the levels, as our level difficulty estimates and ground truth data already have learning "built-in". The DRL agent is trained from scratch for each level, and the human ground truth pass rates are measured from players who have played through the level progression, learning on the way. % . ngry Birds Dream Blast constantly introduces new game elements and mechanics, and the skills learned in early levels only partially transfer to later levels. Our model views skill more of a static trait, although the population skill average does increase over the levels because less skilled players churn earlier.

%TODO: wouldn't it be more logical to increment p.skill instead of s, and draw s,t inside the while loop? Also, why is the boredom random number drawn inside the loop instead of outside the loop. It would look better if the random draws were handled consistently. However, let's not change this if you don't have time to regenerate all results.

\begin{figure}[h]
	\centering
	\includegraphics[width=\linewidth]{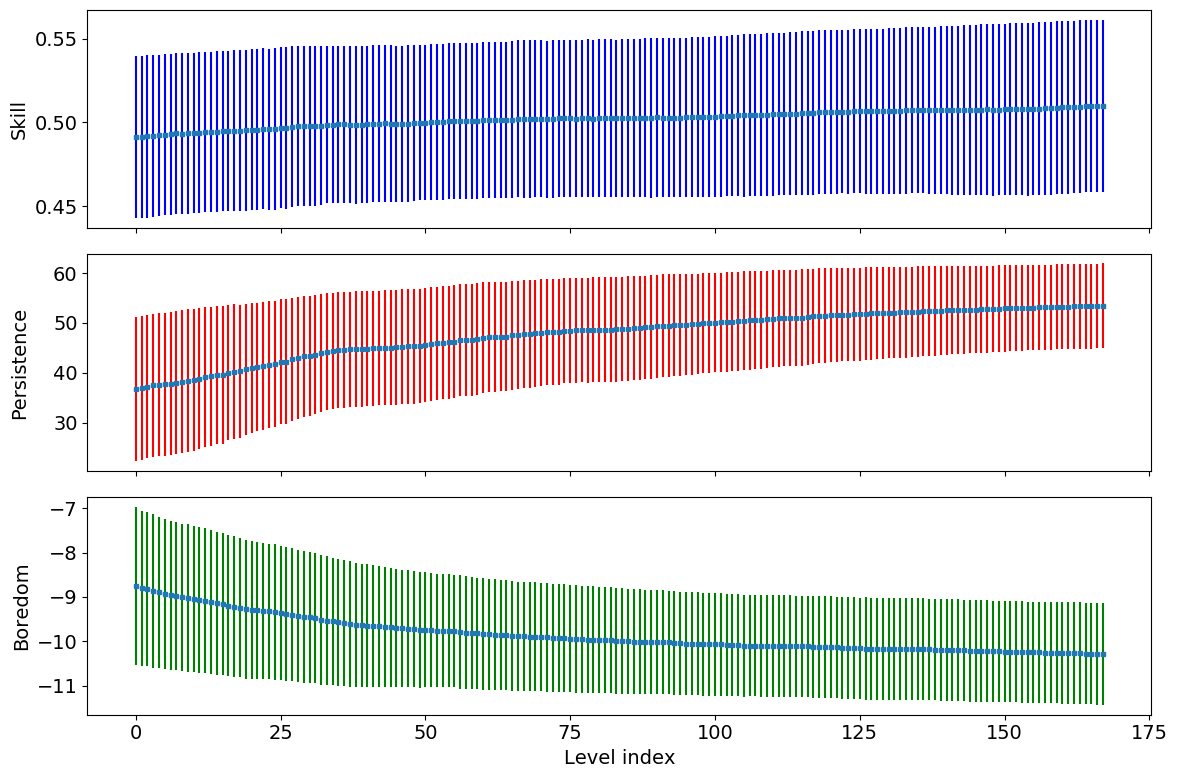}
	\caption{Evolution of the simulated player population's mean and standard deviation of skill, persistence, and boredom attributes over game levels.}
	\label{fig:skill_persistence_bordem}
\end{figure}

Figure \ref{fig:skill_persistence_bordem} shows how skill, persistence, and boredom attribute distributions change over levels in the simulation. One can observe an increase in average persistence, and a decrease in the tendency to get bored. There is also an increase in skill, although to a lesser degree.

\subsubsection{Design Rationale}
We chose to model skill, persistence, and boredom based on our human player data and the literature on game engagement and player psychology. Figure \ref{fig:3_plots} shows that pass rate and churn are related, with players being more likely to churn with low pass rate, in line with \cite{lomas2013optimizing, poromaa2017crushing}. A psychological explanation for this can be found in Self-Determination Theory \cite{ryan2006motivational}, which posits that feelings of competence and mastery support intrinsic motivation. Thus, repeatedly failing a level is likely to decrease the motivation to play. In AB Dream Blast, level pass rate depends on skill but with some randomness due to randomly spawned game objects, which motivates our stochastic skill model. The persistence coefficient is based on some players being more persistent and willing to keep trying when facing hard challenges \cite{neys2014exploring}. Finally, game engagement tends to decline over time \cite{viljanen2016modelling, viljanen2017playtime}. One psychological explanation for this is that the novelty of a game fades over time, which makes players less likely to feel curious interest \cite{silvia2008interest}. Considering boredom as the counterpart of curiosity, we model each simulated player's tendency to get bored as probability of churning after passing a level. This makes the player increasingly likely to churn when progressing through multiple levels and the boredom random draw is repeated.

\subsubsection{Optimizing Simulation Parameters}
To fit the model to observed pass and churn rates from real players, we optimize three types of parameters: 1) the baseline pass rate predictor's regression weights, 2) the means and standard deviations for the initial player population's skill, persistence, and boredom attributes, and 3) the $\alpha,\beta,\theta, \gamma$ parameters of Algorithm \ref{algo:simulation}.

Instead of jointly optimizing all parameters, we simplify by first independently fitting the baseline pass rate predictor parameters. The level difficulty $d$ in Algorithm \ref{algo:simulation} line \ref{line:ifpassed} is computed as $d=normalize(-\rho_b)$, where $\rho_b$ is the baseline pass rate prediction, and the normalization is over all levels, transforming the difficulties between zero and one.

The rest of the parameters (10 in total) are optimized using CMA-ES \cite{hansen2016cma}, a derivative-free global optimization method, to  minimize the following objective function:

\begin{equation}\label{eq:fitness}
f = MSE(\rho_p)+w_\text{churn} MSE(\rho_c),
\end{equation}

where $\rho_p$ and $\rho_c$ are the pass rate and churn rate predictions of Algorithm \ref{algo:simulation}, and $MSE$ denotes computing the mean squared error over all levels. Since the churn rate varies in a much smaller range than pass rate, we employ the $w_\text{churn}$ parameter to amplify the relative imporance of predicting churn correctly. We set $w_\text{churn}$ equal to the variance of human pass rates divided by the variance of human churn rates. 

We use CMA-ES population size of 120 and optimize until there is no improvement for 100 iterations, which takes approximately two hours on an Intel core i7 2.11 GHz CPU and consumes about 170 Mb RAM.
%
%1are computed by applying the negative weights from the first part of the baseline (i.e. pass rate prediction) to the agent's gameplay statistics (Equation \ref{mc_reg1}). We use , an evolutionary algorithm, to minimize the fitness function in equation \ref{eq:fitness}, and find the solution for initial player distribution ($\mu_\text{skill}$, $\sigma_\text{skill}$, $\mu_\text{persistence}$, $\sigma_\text{persistnce}$, $\mu_\text{boredom}$, and $\sigma_\text{boredom}$), and the simulation loop parameters ($\alpha$, $\beta$, $\gamma$, and $\theta$). In total, our optimization problem has 10 dimensions.

%
%\begin{equation}\label{mc_reg1}	
%\mathbf{d} = -\mathcal{X} \mathbf{w}_1 -\mathbf{b}_1,
%\end{equation}
%where $\mathbf{d}$ and $\mathcal{X}$ are level difficulties and agent gameplay statistics, respectively. $\mathbf{w}_1$ and $\mathbf{b_1}$ come from equation \ref{eq:baseline_reg1}.

\begin{algorithm}
	\caption{Pass and churn simulation, run for each level}
\begin{flushleft}
	\textbf{Input}: level difficulty $d$, population $P$, standard deviations $\alpha, \theta, \beta$, learning rate $\gamma$ \\
	\textbf{Output}: pass rate $\rho_p$, churn rate $\rho_c$, evolved population $P$
\end{flushleft}
	\begin{algorithmic}[1]	
		\State $\rho_p \gets 0$	\Comment{Initialization}
		\State $\rho_c \gets 0$	\Comment{Initialization}	
	   \State $M \gets P.size$ \Comment{Remember population size}
		\For{player $p$ in population $P$}
			\State draw $s \sim N(\mu=p.skill, \sigma=\alpha)$ \label{line:s}
			\State draw $t \sim N(\mu=p.persistence, \sigma=\beta)$  \label{line:t}
          \State $passed,churned=$ false
          \State $nAttempts=0$
			\While{\textbf{not} $passed$ \textbf{and} \textbf{not} $churned$}\label{line:while}
				\State $nAttempts \gets nAttempts + 1$
				\If{$s \geq $ level difficulty $d$} \label{line:ifpassed}
					\State $passed \gets \text{true}$
					\State $\rho_p \gets \rho_p + 1/nAttempts/M$
					\State draw $b \sim N(\mu=0, \sigma=\theta)$ \label{line:b}
					\If{$b<p.boredom$} \label{line:ifbored}
						%\Comment{got bored}
						\State $churned \gets \text{true}$
						\State $\rho_c \gets \rho_c+1/M$
						\State $P$.remove($p$)
					\EndIf	
				\Else
					\State $s \gets s+\gamma$ \Comment{Learning} \label{line:learning}
					\If{$nAttempts > t$} \label{line:ifchurn}
						\State $churned \gets \text{true}$
						\State $\rho_c \gets \rho_c+1/M$
						\State $P$.remove($p$)
					\EndIf		
				\EndIf
			\EndWhile
		\EndFor
       \State $P$.add(randomly select $M-P.size$ players from $P$) \label{line:resample}
	\end{algorithmic}
	\label{algo:simulation}
\end{algorithm}
%!TEX root = main.tex
\section{Evaluation}\label{sec:evaluation}
We use 5-fold cross-validation to compute prediction errors and compare the two models. More specifically, we always simulate over all levels, but leave out one fifth in computing the optimization objective function. Thus, human data from levels used for measuring validation error does not inform the model parameter optimization. This matches the intended use case where the model is optimized based on some existing level data, and a designer is using the model to predict pass and churn rates of new levels.

Table \ref{tab:mse_mae} shows the validation mean squared error (MSE) and mean absolute error (MAE) of the two models. The extended model improves churn prediction and performs approximately similarly in pass rate prediction. Note that the two models are complementary instead of mutually exclusive: As the baseline pass rate predictor gives a slightly lower mean error, one might use its predictions as such, and only utilize the churn prediction of the extended model. The means and standard deviations are computed over the cross-validation folds. As CMA-ES results vary with random seed, we ran the cross-validation 5 times for the extended model, i.e., the mean and standard deviation are over a total of 25 optimization runs. 
\begin{table*}[ht]
	\centering
	\caption{Mean squared errors and mean absolute errors of average pass rate and churn rate prediction.}
	\label{tab:mse_mae}
	\begin{tabular}{|c|p{0.135\linewidth}|p{0.135\linewidth}|p{0.135\linewidth}|p{0.135\linewidth}|}
		\hline	
		& \multicolumn{2}{|c|}{Validation MSE}& \multicolumn{2}{|c|}{Validation MAE} \\ \cline{2-5}
		Method&Pass rate &Churn rate &Pass rate&Churn rate \\ \hline
		Baseline&\makecell{$\mu=0.02244$ \\ $\sigma=0.00803$}&\makecell{$\mu=0.00013$ \\ $\sigma=0.00003$}&\makecell{$ \mu=0.11228$\\ $\sigma=0.01663$}&\makecell{$\mu=0.00866$ \\ $\sigma=0.00076$}\\ \hline
		
		Extended model &\makecell{$\mu=0.02320$ \\ $\sigma=0.00831$}&\makecell{$\mu=0.00008$ \\ $\sigma=0.00002$}&\makecell{$ \mu=0.11467$\\ $\sigma=0.01647$}&\makecell{$\mu=0.00607$ \\ $\sigma=0.00073$}\\ \hline
		%Baseline &\makecell{0.02244}&\makecell{0.00013}&\makecell{0.11228}&\makecell{0.00866}\\ \hline
	    %Extended model &\makecell{$\mu=0.02320$ \\ $\sigma=0.00019$}&\makecell{$\mu=0.00008$ \\ $\sigma=0.00003$}&\makecell{$ \mu=0.11467$\\ $\sigma=0.00044$}&\makecell{$\mu=0.00607$ \\ $\sigma=0.00022$}\\ \hline
	\end{tabular}
\end{table*}

\begin{figure}[h]
	\centering
	\includegraphics[width=\linewidth]{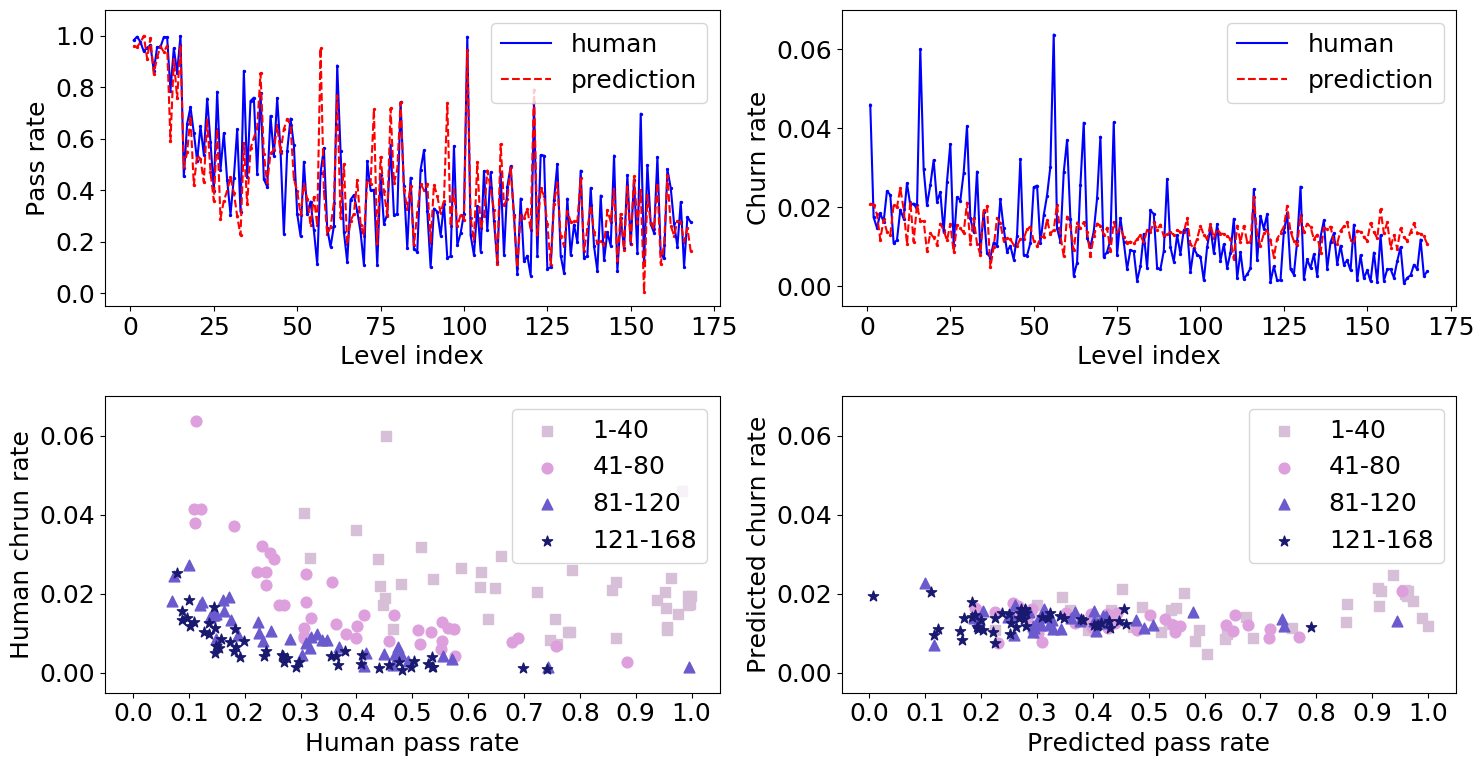}
	\caption{Prediction results of the baseline model. Scatter plot colors correspond to game level indices.}
	\label{fig:baseline_plot}
\end{figure}

\begin{figure}[h]
	\centering
	\includegraphics[width=\linewidth]{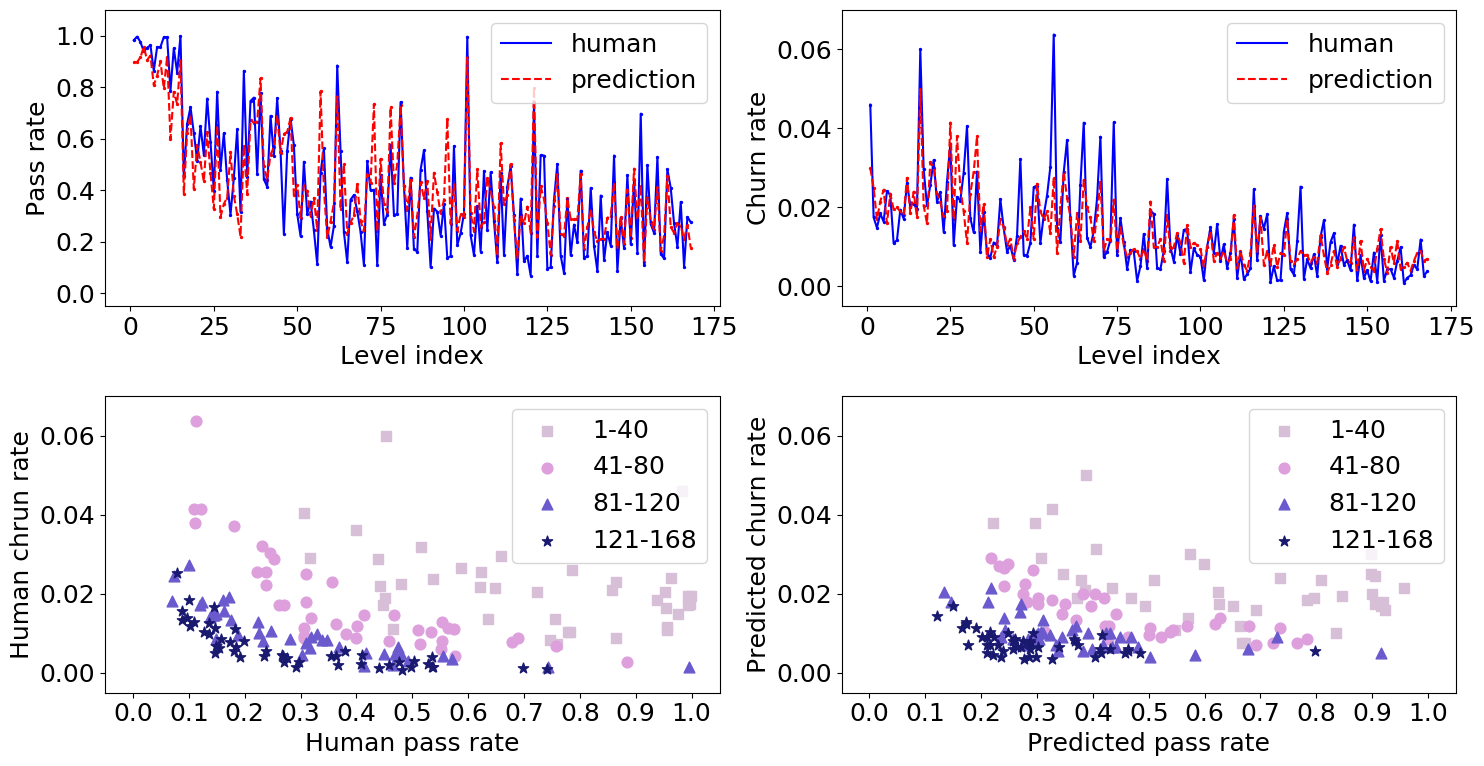}
	\caption{Prediction results of our extended model. Scatter plot colors correspond to game level indices.}
	\label{fig:modelV3_plot}
\end{figure}

%We used the best model out of 5 cross validation %runs to create the figures \ref{fig:baseline_plot} %and \ref{fig:modelV3_plot}.
As shown in the scatter plots in Figures \ref{fig:baseline_plot} and \ref{fig:modelV3_plot}, the extended model is better able to discriminate between the early and later game levels. As players with low skill and persistence churn earlier, same as players who get bored easier, the average churn rate in early levels is higher than in the later levels. 

To test the generalization to new levels added to the end of the game, we did an extra validation test that leaves out the last fifth of the levels during training. The pass and churn rate MSEs of these levels are 0.01003 and 0.00003, which are lower than the baseline 0.01073 and 0.00016. Naturally, our model can become less reliable when more new levels are added, especially if the levels feature novel gameplay. Thus, when releasing new levels, a game developer should collect human player data and refit the population simulation parameters.
%Figure \ref{fig:skill_persistence_bordem} has been illustrated before adding noise to the skill and persistence of the players (line 3 and 4 , Algorithm \ref{algo:simulation}). However, to get the good solution from the optimization, adding noise is required; and it indicates that even the mean of players skill and persistence increase over level funnel, we need different variety of players (i.e. high variance). Probably, The reason is that some players pass the early levels by chance.

%PERTTU: I commented the following out as it may be confusing. We may of course add it back in if reviewers ask about it
%In the case of considering just the average churn ratio after failure, the baseline can predict churn ratio better, but still not good as our model (Figure \ref{fig:baseline_mc_failure}). 
%\begin{figure}[h]
%	\centering
%	\includegraphics[width=\linewidth]{figures/baseline_mc_failure.png}
%	\caption{Churn ratio after failure prediction. a. baseline. b. our model.}
%	\label{fig:baseline_mc_failure}
%\end{figure}

\begin{figure}[h]
	\centering
	\includegraphics[width=\linewidth]{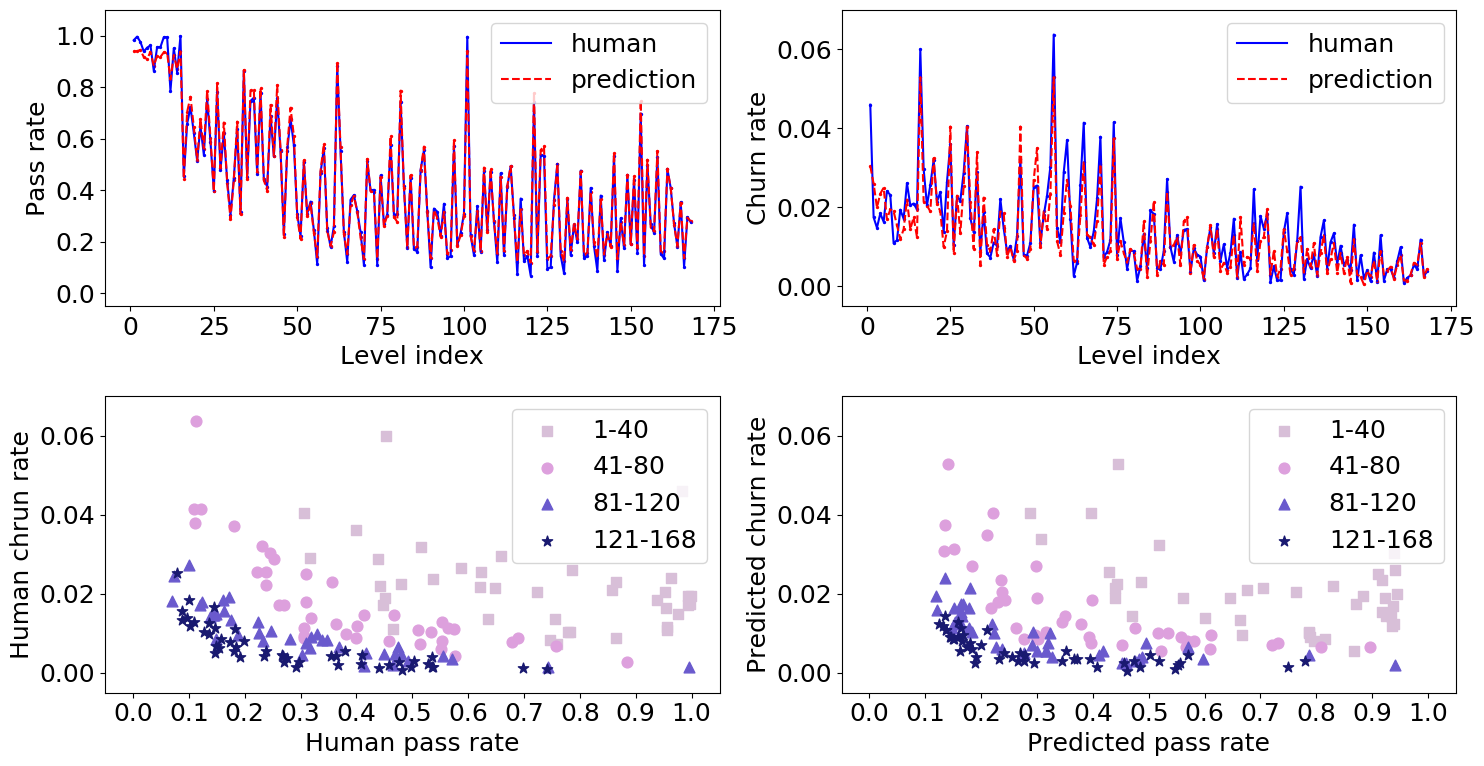}
	\caption{Extended model results when computing level difficulty based on actual human pass rates instead of those predicted by AI gameplay. Comparing this to Figure \ref{fig:modelV3_plot}, one notes improved churn prediction for low pass rates.  Scatter plot colors correspond to game level indices.}
	\label{fig:limitation}
\end{figure}

When carefully comparing the predicted and ground truth data in Figure \ref{fig:modelV3_plot}, one notes that although our model captures the overall relation, some inaccuracy remains especially at low pass rates. We hypothesized that this is at least to some degree due to inaccuracies of the pass rates predicted by the AI gameplay. We tested this hypothesis by also running the simulation using $d=1-\rho_\text{human}$ as the difficulty estimate, where $\rho_\text{human}$ is the human pass rate. This decreases churn prediction MSE by 71\%. Figure \ref{fig:limitation} also shows how the scatter plots are visually more accurate. While the results indicate our AI game playing agents are not yet human-like enough, the results also suggest that our pass and churn simulation of the extended model is surprisingly accurate.

We performed an ablation study to investigate the effect of each simulation feature on the prediction results. Table \ref{tab:mse} indicates that removing any of the features would make our churn model less accurate.
\begin{table*}[ht]
	\centering
	\caption{The effect of different simulation components on the prediction results.}
	\label{tab:mse}
	\begin{tabular}{|c|c|c|}
		\hline
		Method&Pass rate validation MSE&Churn rate validation MSE \\ \hline
		All features&0.023&0.00008 \\ \hline
		No boredom&0.024&0.00009 \\ \hline
		No persistence&0.023&0.00010 \\ \hline
		No learning&0.032&0.00010 \\ \hline
		No random noise in skill and persistence&0.023&0.00014 \\ \hline
	\end{tabular}
\end{table*}


%One limitation of our work is that if there is enormous error between level difficulty estimated by the agent and the actual difficulty of the level, it reduces the prediction accuracy for all the subsequent levels since the player population passed to the following levels is not reliable anymore. \textcolor{red}{Using a linear regression to learn the player population parameters (mean, std) for each level would alleviate this problem. Instead, we can add parameter correction by start the simulation loop from the specific level not just the first level.}
%
%You can see in Figure \ref{fig:limitation}, when we compute the level difficulty as one minus actual average human pass rates, then churn prediction MSE would decrease about $0.00005$.

%\input{future_work.tex}
%!TEX root = main.tex
\section{Limitations and Future Work}
As a limitation, our model has not yet been validated in actual game design work, e.g., in testing new levels and identifying problematic levels before they are deployed to real players. This should be possible in future work, as our Deep Reinforcement Learning game playing approach does not require any reference gameplay data and should thus generalize to new levels. Our cross-validation also suggests generalizability. As a downside, we noticed that the difficulty predictions of the DRL data are not yet entirely human-like especially for levels with very low pass rates. This is a topic to investigate in future work, especially as the accuracy of our population simulation and churn prediction does increase considerably when replacing DRL difficulty predictions with ones computed from ground truth human data. 

Training a DRL agent on a new level is also slow, which imposes a bottleneck on simulated game testing efficiency. Ways to address this in future work include improving parallelism (e.g., \cite{adamski2018distributed}) and testing and comparing alternative game playing AI approaches, e.g., utilizing MCTS instead of DRL, or speeding up the DRL training process using transfer learning. 

\section{Conclusion}\label{sec:conclusion}
We have proposed a novel simulation model for predicting puzzle game level pass and churn rates. Our model combines both AI gameplay for level difficulty prediction, and simulating the evolution of a player population over game levels using simple computational models of skill, persistence, boredom, and learning. In terms of cross-validation error, the model outperforms a baseline model that predicts pass rates and churn based on AI gameplay data alone. %, which gives evidence of the value of the population simulation. 

Our work shows how predictions produced by game playing Deep Reinforcement Learning agents can be enhanced by even a computationally very simple population-level simulation of individual player differences, without requiring retraining the agents or collecting new gameplay data for each simulated player. As training the DRL agents can be very slow, this is a foundational result that can empower future research and applications of player modeling and simulation-based game testing. Our model is also the first to make human-like pass and churn rate data emerge from AI gameplay in a way that captures how the relation of pass rate and churn evolves over a game's level progression.

\section{Acknowledgments}\label{sec:acknowledge}
We thank the anonymous reviewers for their thorough and constructive comments.

%%
%% The next two lines define the bibliography style to be used, and
%% the bibliography file.
\bibliographystyle{ACM-Reference-Format}
\bibliography{references}

%%
%% If your work has an appendix, this is the place to put it.
\end{document}